# The Improvement of Negative Sentences Translation in English-to-Korean Machine Translation

Jang Chung Hyok

Foreign Language Faculty

**Kim Il Sung** University

Pyongyang, Democratic People's Republic of Korea

Kim Kwang Hyok

Computer Science College

**Kim Il Sung** University

**Abstract**

This paper describes the algorithm for translating English negative sentences into Korean in English-Korean Machine Translation (EKMT). The proposed algorithm is based on the comparative study of English and Korean negative sentences. The earlier translation software cannot translate English negative sentences into accurate Korean equivalents. We established a new algorithm for the negative sentence translation and evaluated it.

**Introduction**

Translating English negative sentences into Korean is fairly challenging since the ways of expressing negation in the two languages are not exactly the same. The generally accepted way of the negative sentence translation in EKMT was to simply negate the predicate of its corresponding affirmative sentence, which cannot give the accurate Korean equivalent for some English negative sentences, e.g. partial negatives. We did linguistic comparison between the ways of expressing negation in English and Korean and propose an algorithm for the accurate Korean equivalents to the English negative sentences in EKMT.

1. **Linguistic comparison of English and Korean negation**

Linguistically, negation refers to a grammatical construction with negative polarity [1]. The negative sentence and its affirmative correspondent are closely related in English and so are they in Korean. This requires that we should consider the English negative sentences in close relation with their affirmative correspondents and Korean equivalents.

English and Korean negative sentences have the following features in common;

a. The negative sentences are declaratives, interrogatives or imperatives in both languages.
b. The negative sentences have their affirmative correspondents, which can be made negative by adding negative words.

c. Negative words are relatively restricted in number.
d. In some cases, the negative words are used with other negation-related words to intensify or soften the negation.
e. Negation is subdivided into general negation, partial negation, double negation and intensified negation.

Despite these common features, there are still differences, the most important one of which is the way of adding negative words to the affirmative sentences. The table below gives the comparison between English and Korean negative sentences. The underlined words are the negative words and the bold words are the negation-related words in English and Korean.

| Part of Speech | Negative Words | English Negative Sentences | Korean Equivalents | Corresponding Affirmative Sentences |
| --- | --- | --- | --- | --- |
| Adverb | not | I did not believe **at all** that he was coming. | 나는 그가 온다는것을 **전혀** 믿지 않았다. | I believed that he was coming. |
| | | Not much is known about the disease. | 그 질병에 대하여 많은것이 알려지지는 않았다. | Much is known about the disease. |
| | | Not many people have read the report. | 많은 사람들이 그 기사를 읽은것은 아니다. | Many people have read the report. |
| | | Not one of the students could answer. | 학생들가운데서 **한사람도** 대답할수 없었다. | One of the students could answer. |
| | never | He has never been there. | 그는 거기에 가본적이 **전혀** 없다. | He has been there. |
| | | Never again would he say that. | 그는 다시는 그런 말을 하지 않을것이다. | He would say that again. |
| | hardly | I can hardly believe it. | 나는 그것을 **거의나** 믿을수 없다. | I can believe it. |
| | little | Little did I know that my life was about to change. | 나는 나의 생활이 변하리라는것을 **거의나** 알지 못하였다. | I knew that my life was about to change. |
| | nowhere | He has nowhere to live. I am going nowhere. | 그는 살곳이 없다. 나는 **아무데도** 가지 않는다. | He has somewhere to live. |
| | nothing | She's nothing like her brother. | 그는 자기의 동생과 **전혀** 비슷하지 않다. | She's like her brother. |
| | none | She seems <u>none</u> the worse for her experience. | 그는 자기 경험에 비하여 한심해보이<u>지는 않는다</u>. | She seems the worse for her experience. |
| Pronoun | no one | <u>No one</u> could hear me. | **그 누구도** 나의 말을 듣<u>지 못하였다</u>. | Someone could hear me. |
| | nobody | I knocked on the door but <u>nobody</u> answered. | 내가 문을 두드렸지만 **그 누구도** 대답하<u>지 않았다</u>. | I knocked on the door and somebody answered. |
| | nothing | We said <u>nothing</u> about her. | 우리는 그에 대하여 **아무것도** 말하<u>지 않았다</u>. | We said something about her. |
| | none | <u>None</u> of my friends phones me any more. | 내 친구들중 **그 누구도** 나에게 더 이상 전화하<u>지 않는다</u>. | Some of my friends phone me. |
| | little | <u>Little</u> is known about the causes of the problem. | 그 문제의 원인에 대해서는 **거의** 알려지지 않았다. | Something is known about the causes of the problem. |
| Determiner | no | <u>No</u> food is left in the fridge. | 랭동기에는 음식이 남은것이 **하나도** <u>없다</u>. | Some food is left in the fridge |
| | few | <u>Few</u> people knew he was ill. | 그가 앓는다는것을 아는 사람은 **거의** 없다. | Some people knew he was ill. |
| | little | I paid <u>little</u> attention to what the others were saying. | 나는 다른 사람들이 하는 말에 **거의** 관심을 돌리<u>지 않았다</u>. | I paid attention to what the others were saying. |

Table 1. Comparison between English and Korean negative sentences

As the table shows the negative words in English include adverbs, pronouns and determiners, therefore the subject, predicate, object, adverbial modifiers or attributes can be negated, while in Korean the negation is expressed by adding agglutinative suffixes (the underlined words in Korean sentences) to the predicates and by adding other related words (the bold words in Korean sentences) to the sentences in some cases. This significant difference poses a main problem for the translation of the English negative sentences into Korean.

2. **Classification of English negative sentences**

Some English sentences with negative words are not negative sentences, so they are not translated into Korean negative sentences.

*He just stopped working for <u>no</u> reason.* (그는 아무 리유도 없이 일을 그만두었다.)
*The victory was <u>nothing</u> less than a miracle.* (그 승리는 그야말로 기적이였다.)

Also, the following examples show that there is a big difference in word order between the negative sentence and its corresponding affirmative sentence in English, while their Korean equivalents have less difference except the negative words and the negation-related words.

*Little did I know that my life was about to change.*
(나는 나의 생활이 변하리라는것을 **거의나** 알지 못하였다.)
*I knew that my life was about to change.*
(나는 나의 생활이 변하리라는것을 알고있었다.)

Considering the features described above, we concluded that it is very important to decide whether the sentence with a negative word is a negative sentence and what sort of negative sentence it is.

a. If the subject, predicate or the object of the sentence is negated or a negative word itself, it is a negative sentence.

*No one could hear me.* (negative subject)
*We said nothing about her.* (negative object)
*He has never been there.* (negative predicate)

However, if the negative word is not the subject, object or predicate of a sentence but a part of an idiomatic expression (e.g. for no reason, be nothing less than, etc), the Korean equivalent might not be a negative sentence.

b. We subdivided the English negative sentences into single negative (one negative word in the sentence) and double negative sentences (two or more negative words in the sentence) according to the number of negative words in the sentence, and also into general negative, partial negative and intensified negative sentences according to the part of sentence which is negated.

The Korean equivalent of the general negative sentence can be made from its corresponding affirmative sentence by adding agglutinative suffix, while those of the partial negative and intensified negative sentences require other negation-related words as well.

### 3. Translation of English negative sentences

3.1 The determination of English negative sentence

To translate English negative sentence, the parser should decide;

a. whether the sentence with a negative word is a negative sentence.

According to which part of the sentence is negated, we subdivided the negative sentences into six categories;

- Negative Subject – Affirmative Predicate – Affirmative Object (NS-AP-AO)
- Negative Subject – Negative Predicate – Affirmative Object (NS-NP-AO)

- Negative Subject – Affirmative Predicate – Negative Object (NS-AP-NO)
- Affirmative Subject – Affirmative Predicate – Negative Object (AS-AP-NO)
- Affirmative Subject – Negative Predicate – Affirmative Object (AS-NP-AO)
- Affirmative Subject – Negative Predicate – Negative Object (AS-NP-NO)

If the sentence matches one of the above sentence structures, the parser decides that it is a negative sentence.

b. what sort of negative sentence it is.

According to the number of negative words and the negated part of sentence we conclude that;

- Single negative sentence is general negative, partial negative or intensified negative.

The negative sentence with negative subject, negative object or negative predicate which is modified by the words like 'every, all, always, entirely, necessarily, etc' is a partial negative sentence.

- Single negative sentence with negative predicate which is not a partial negative sentence is a general negative sentence.
- Single negative sentence with negative subject or negative object which is not a partial negative sentence is an intensified negative sentence.
- Double negative sentence is an intensified negative sentence.

c. which negative agglutinative suffixes and related words should be added to the corresponding Korean affirmative sentence.

Korean language has various negative agglutinative suffixes to the predicate such as "~이 아니다", "~지 못하다", "~지 않다" and "~지 말다". It is not easy to decide which one makes a natural expression in Korean. In 3.2 and 3.3, we describe the principles to choose correct suffixes and related words to the corresponding Korean affirmative sentences.

3.2 Negative agglutinative suffixes to Korean predicate

The negative agglutinative suffixes to Korean predicate is decided by;

a. The kind of predicate verb in English sentence and its tense and aspect

If the predicate is "a copular verb + noun", add "~이 아니다" or "~이 없다" to the predicate of the Korean affirmative sentence, and if the predicate is a lexical verb and the auxiliary verb *do* is used, add "~지 않다" or "~지 못하다".

*She is not my younger sister.*
그 녀자는 나의 녀동생이 아니다.

*He doesn't get up early in the morning.*

그는 아침 일찍 일어나<u>지 않는다</u>.

If the predicate is a lexical verb in perfect aspect, add "~ㄴ적이 없다".

*He has never been there.*

그는 거기에 가<u>본적이 없다</u>.

b. The kind of English negative sentence

If the negative sentence is a general negative, the above rule is applied to choose a suffix to Korean predicate.

If it is an intensified negative, the Korean negative suffix is added as general negative sentence.

*No trains will be affected by this accident.*

렬차들은 이 사고에 의하여 전혀 영향을 받<u>지 않을것이다</u>.

*He has made promises, but kept none of them.*

그는 약속을 하였지만 하나도 지키<u>지 않았다</u>.

If it is a partial negative, the agglutinative suffix "~ㄴ 것은 아니다" is added.

*It is not always true that combining two substances into one releases energy.*

두 물질을 하나로 결합하면 에네르기가 방출된다는것이 언제나 진실<u>인것은 아니다</u>.

If an auxiliary verb is negated in a partial negative sentence, the suffixes "~지는 않다" or "~수는 없다" is added to the Korean equivalent of the lexical verb and change it according to the tense of the English predicate.

*The particles will not necessarily move the same way under identical conditions.*

립자들이 동일한 조건에서 반드시 같은 방식으로 움직이<u>지는 않을것이다</u>. ○
립자들이 동일한 조언에서 반드시 같은 방식으로 움직<u>일것이지는 않다</u>. ×

c. The collocation of the Korean predicate

In Korean, the terminative predicates of sentences are verbs or adjectives. According to the characteristics of Korean verb or adjective collocation, the suffix "~지 않다" or "~지 못하다" is added.

*Their effort accomplished little or nothing.*

그들의 노력은 거의 아무것도 달성하<u>지 못하였다</u>. ○
그들의 노력은 거의 아무것도 달성하<u>지 않았다</u>. ×

The Korean equivalents of some English negative predicates are completely different from their Korean affirmatives, thus making them impossible to be negated by adding agglutinative suffixes. In this case, the Korean predicate is replaced by appropriate word.

*I don't know what these symbols mean.*

나는 이 기호들이 무엇을 의미하는지 <u>모른다</u>.

3.3 Negation-related words

Some negatives sentences (e.g. intensified negatives and double negatives) are accurately translated in Korean not only by adding negative suffix to the Korean predicate but also by adding adverb and changing the particle to the subject or the object.

  a. If it is a single negative (one negative word in the sentence), the negation-related words are added according to the English negative word.
  - If the predicate verb is negated by adverb "not", no related words are added.
    *The rooms don't have air-conditioners.*
    방들에는 공기조화기가 없다.
  - If the negative word in the sentence is "never", "little" or "few", the related adverbs are added to the Korean equivalent.
    *She never looked like her photograph.*
    그 녀자는 자기 사진과 <u>전혀</u> 비슷해보이지 않는다.
  b. If it is a partial negative, the particle added to the subject of a Korean equivalent changes from 는(은) to 가(이).
    *Not all my friends were at the wedding.*
    나의 모든 친구들<u>이</u> 결혼식에 참가한것은 아니였다.
  c. If it is an intensified negative in which the subject or the object is negated, the particle to the Korean subject or object noun should be changed or other related words added.
  - If the subject or object is a negative pronoun or is negated by the determiner "no", the particle to the Korean subject or object noun should be changed.
    *Not one person remembered my birthday.*
    한 사람<u>도</u> 나의 생일을 기억하지 못하였다.
    *I have seen none of them before.*
    나는 그들중 누구<u>도</u> 이전에 본적이 없다.
  - If the subject or object is negated by the determiner "little" or "few" to make a quasi-negative, negation-related words are added.
    *Few students could answer the question.*
    <u>거의</u> 모든 학생들이 그 질문에 대답할수 없었다.
    *He showed little interest in my suggestion.*
    그는 나의 제안에 <u>거의</u> 흥미를 보이지 않았다.
  d. If it is a double negative, the particles to the subject or the object should be changed and add negation-related words as the intensified negative.

*I never said nothing about the plan.*

나는 그 계획에 대하여 아무것도 말하지 않았다.

## 4. The improved algorithm for the English negative sentences

On the basis of the linguistic study described above, we established an improved algorithm for the English negative sentence translation in EKMT. The following diagram shows it.

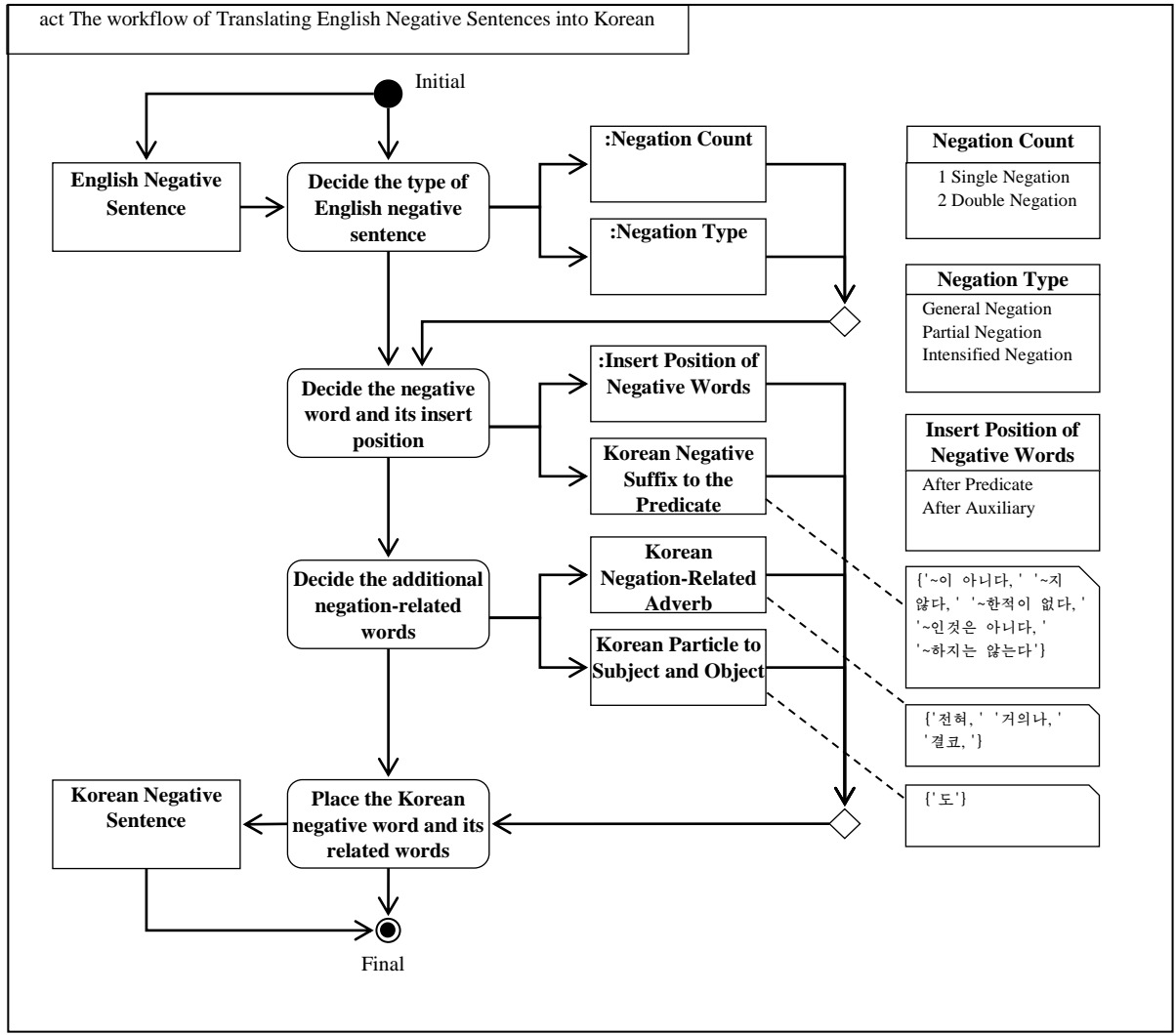

Fig.1. Algorithm for the English negative sentence translation

## 5. Evaluation

We applied the improved algorithm to English-to-Korean Translation System "Ryongnamsan" to improve its translation accuracy. The sample text was about the computer operating system. The sentences with negative words are 518 or 8.2% of the whole 6377 sentences. The following table shows the number of negative sentences of each kind.

| Single Negative | | | Double Negative | | | Others |
|---|---|---|---|---|---|---|
| NS-AP-AO | AS-NP-AO | AS-AP-NO | NS-NP-AO | NS-AP-NO | AS-NP-NO | |
| 56 | 408 | 41 | 0 | 0 | 0 | 13 |

Table 2. Number of negative sentences of each kind

There were 5 partial negative sentences among 505 single negatives. The following examples show the translation results of the partial negative and intensified negative by the proposed algorithm, in comparison with the earlier one.

*When a process is executed, it is not always executed at the site in which it is initiated.*
프로쎄스가 실행될 때 그것은 언제나 호출이 시작된 싸이트에서 실행되지 않는다. (by the earlier algorithm)
프로쎄스가 실행될 때 그것이 언제나 호출이 시작된 싸이트에서 실행되는것은 아니다. (by the proposed algorithm)

*Servers cannot do so, since they have no knowledge of the purpose of the client's requests.*
그것들이 의뢰기의 요청들에 대한 지식을 가지지 않기때문에 봉사기들은 그렇게 할수 없다. (by the earlier algorithm)
그것들이 의뢰기의 요청들에 대한 그 어떤 지식도 가지지 않기때문에 봉사기들은 그렇게 할수 없다. (by the proposed algorithm)

As the examples show, the proposed algorithm improved the translation of negative sentences, making it accurate and similar to the human translation.

## 6. Conclusion

Negative sentences don't account for large proportion in English text, but their translation is important. We did a comparative study on the negation in English and Korean to establish a new translation algorithm in EKMT and introduced it to English-to-Korean Machine Translation System "Ryongnamsan" and conclude that it is effective.